%% file: main.tex
\documentclass{article} 
\usepackage{iclr2027_conference,times}

\input{math_commands.tex}

\usepackage{wrapfig}
\usepackage{graphicx}
\usepackage{subcaption}
\usepackage{enumitem}
\usepackage{tabularx}
\usepackage{booktabs}
\usepackage{multicol}
\usepackage{multirow}
\usepackage[export]{adjustbox}
\usepackage{listings}

\usepackage[normalem]{ulem}
\usepackage[most]{tcolorbox}
\definecolor{csaccent}{RGB}{24,68,112}
\definecolor{csold}{RGB}{145,52,52}
\definecolor{csnew}{RGB}{0,86,140}
\definecolor{csresp}{RGB}{245,247,249}
\newcommand{\oldobj}[1]{\textcolor{csold}{\sout{#1}}}
\newcommand{\newobj}[1]{\textcolor{csnew}{\textbf{#1}}}
\newcommand{\swap}[2]{\mbox{\oldobj{#1}\,\textcolor{black!45}{$\rightarrow$}\,\newobj{#2}}}

\usepackage{xcolor}
\usepackage{hyperref}
\usepackage{url}

\definecolor{linkblue}{HTML}{1565C0}

\hypersetup{
  colorlinks  = true,
  linkcolor   = black,     
  citecolor   = linkblue,  
  urlcolor    = linkblue,  
  filecolor   = linkblue,
}

\setlist[itemize]{
    leftmargin=1.5em,  
    rightmargin=0pt, 
    itemsep=0.25em,     
    topsep=0em       
}

\title{Generalizable Model Editing Could Be More Than Direct Generalization}

\author{Xiaotian Ye\textsuperscript{1}\thanks{Equal contribution.}\ , Chunyao Yang\textsuperscript{2,3}$^*$, Xiaohan Wang\textsuperscript{4}, Junkai Chen\textsuperscript{2}, \\ \vspace{0.5em}
\textbf{Yongqi Duan}\textsuperscript{5}, \textbf{Mengqi Zhang}\textsuperscript{6}, \textbf{Shu Wu}\textsuperscript{2} \\
\textsuperscript{1}University of Cambridge ~ 
\textsuperscript{2}NLPR, MAIS, Institute of Automation, Chinese Academy of Sciences \\
\textsuperscript{3}School of Advanced Interdisciplinary Sciences, University of the Chinese Academy of Sciences \\
\textsuperscript{4}HUST ~ \textsuperscript{5}BUPT ~\textsuperscript{6}Shandong University
\\
\texttt{xy401@cam.ac.uk, \{yangchunyao2025, chenjunkai2026\}@ia.ac.cn}\\
\texttt{shawn\_wang@hust.edu.cn, duanyongqi@bupt.edu.cn}\\
\texttt{mengqi.zhang@sdu.edu.cn, shu.wu@nlpr.ia.ac.cn}
}

\iclrfinalcopy 
\begin{document}

\maketitle

\begin{abstract}
A central desideratum in model editing is generalization to complex tasks, which prior work formalizes under a \textit{direct generalization} paradigm: treating and evaluating LLM behavior as an input-output mapping and expecting a factual update to alter outputs on reasoning tasks. Yet despite substantial empirical effort, the field consensus remains that such generalization is highly challenging and elusive. However, when we deploy the latest models in an in-the-wild chatbot-user setting, we surprisingly find that even classical editing methods exhibit remarkably fluent generalization behavior. This motivates us to revisit the obstacles to generalization and the limitations of the prevailing paradigm. We find that direct generalization is fundamentally constrained both theoretically and practically: theoretically, it suffers from an identifiability issue; and practically, existing approaches attempting to circumvent this are further hindered by the ambiguity of their constructed biases and by vulnerability to being overridden by other factors such as optimization dynamics. We instead identify \textit{compositional generalization}, empowered by the increasingly strong reasoning and composition abilities of modern LLMs, as a theoretically grounded and empirically validated alternative. We validate this phenomenon across multiple models and editing methods, showing that such generalization behavior is widespread. Overall, we argue that generalizable model editing could be more than direct generalization, and call for a paradigm shift toward more generalizable and practically deployable knowledge update techniques.
\end{abstract}

\section{Introduction}
\label{sec:intro}

The rapid rise of LLMs has reshaped the landscape of AI. However, the internal knowledge stored within these models inevitably suffers from obsolescence, hallucinations, and factual errors \citep{Ji2023hallusurvey,lin-etal-2022-truthfulqa}. To precisely rectify or update specific factual knowledge in LLMs without full-scale training, Model Editing has emerged as a promising paradigm \citep{zhang2024comprehensivestudyknowledgeediting,yao-etal-2023-editing,zhang-etal-2026-spectral}, whose objective is to modify specific facts in an efficient and low-resource manner. 
Specifically, an ideal editing method is expected to satisfy three core desiderata: \textit{Efficacy}, where the model accurately answers the targeted knowledge; \textit{Locality}, which ensures that unrelated knowledge and capabilities remain intact; and \textit{Generalization}, which denotes the capacity not merely to memorize given samples, but to extrapolate the knowledge to unseen complex tasks \citep{yao-etal-2023-editing}.

After years of development, the generalization in model editing remains far from satisfactory. Concretely, current mainstream practice can be summarized as a \textbf{direct generalization} paradigm: taking generalization on multi-hop reasoning as an example, given an edit $e_1=(s_1,r_1,o_1^*)$, mainstream evaluations construct compound question prefixes that have never appeared during editing or training \citep{zhong-etal-2023-mquake,zhang2024comprehensivestudyknowledgeediting}; for example, constructing a multi-hop question $(s_1, r_1, r_2)$ and then examining whether the model produces the target answer $o_2^*$ as a direct continuation of that prefix. This essentially treats the model as an input-output mapping, and formalizes generalization as whether the edit can encode the inferential consequence implied by the edit into a new input-output rule. However, the field consensus is that the generalization ability of editing is extremely poor: prior work has shown that overfitting is more commonly observed \citep{zhang2025uncovering,cohen2024evaluating}, with little to no fluent and natural generalization behavior, severely limiting the practical applicability of model editing in real-world deployment.

In this paper, however, we for the first time observe an unexpected generalization behavior in a more in-the-wild setting: we apply several classic conventional model editing methods to the latest models and deploy them in the most common chatbot-user interaction manner. Upon querying complex questions, without any additional prompt engineering or structural modifications, we observe that a high proportion of cases exhibit remarkably strong generalization behavior. This motivates us to question whether generalization in model editing is truly as challenging as believed, or whether the observed difficulty arises from fundamental issues inherent in the previous paradigm? In this paper, we conduct a deep investigation into this question.

First, we find that the pervasive failure of generalization is highly correlated with limitations of the paradigm itself, including its implicit evaluation protocol, at both theoretical and practical levels. Theoretically, we introduce the perspective of \textit{identifiability} in autoregressive models: 
in the editing setting, the completion distribution over out-of-distribution (OOD) and complex tasks is unidentifiable from the edit data alone. Consequently, generalization reduces to selecting a correct model solution from an unidentifiable set of candidate models that all fit the target edit distribution—a process fundamentally governed by inductive biases, which current paradigms leave entirely to the editing algorithm despite these biases remaining poorly understood. From a practical perspective, we discuss existing approaches to achieving generalization and conduct a deeper analysis of the mainstream locate-then-edit approach. We find that this line of methods has long been hindered by two severe obstacles in practice: while such methods can be viewed as manually constructing better inductive biases grounded in interpretability, we find that this endeavor is plagued by the inherent ambiguity of the constructed biases, and that these biases may further be overridden by other biases such as optimization dynamics and engineering factors. We discuss the severity of these issues through two case studies and construct a baseline that adjusts only a few hyperparameters to demonstrate their practical impact on performance; this simple baseline successfully outperforms the established method MEMIT-LTI \citep{zhang2025uncovering}, which employs carefully constructed constraints to mitigate overfitting. These findings jointly explain why direct generalization has long proved genuinely elusive in the literature.

A major trend in LLMs has been the continuous enhancement of reasoning capabilities, which makes models exhibit increasingly powerful abilities to combine different atomic knowledge and capabilities to solve complex problems \citep{wei2022emergent}. We propose a new perspective: this inherent meta-ability of composition acts as a powerful structural prior, serving as a generalized, implicit inductive bias for the editing process. This theoretically enables the \textit{compositional generalization} of model editing: if a model possesses the ability to compose the edit with other knowledge and capabilities, generalization need not be accomplished entirely by the editing algorithm but can emerge naturally from the compositional ability. Yet this very capability has been inadvertently suppressed by prior evaluation settings, as the RL-based training and capability enhancement of modern LLMs are built upon chat models \citep{openai2023gpt,yang2025qwen3technicalreport} rather than the unstructured text completion that previous work focused on. Our experiments demonstrate that methods that overfit under conventional settings can exhibit highly fluent generalization behavior in the wild, with generalization performance strongly correlated with the inherent capabilities of the model. Extensive analyses further validate the promising potential of this paradigm, and we therefore argue that \textit{Generalizable Model Editing Could Be More Than Direct Generalization}. Overall, our contributions are:

\begin{itemize}
\item We highlight the limitations of the current direct generalization paradigm and analyze its fundamental obstacles grounded in identifiability perspective.
\item We conduct a systematic analysis of existing editing methods based on the above theoretical framework, further identifying several key obstacles faced by mainstream approaches.
\item We introduce and validate \textit{compositional generalization} as a new perspective, demonstrating across multiple models and editing methods that this phenomenon is widespread.
\end{itemize}

\section{Related Works}

In this section, we briefly review the related work on knowledge editing, with more detailed related work and technical backgrounds provided in Appendix~\ref{app:related_work} and~\ref{app:background}.

Knowledge editing aims to precisely and efficiently update the internal knowledge of LLMs \citep{10.1145/3698590,zhang2024comprehensivestudyknowledgeediting}, in the mainstream setting each piece of knowledge is formulated as a triple $(s, r, o)$ \citep{de-cao-etal-2021-editing}, consisting of a subject $s$, relation $r$, and object $o$. Mainstream approaches can be broadly categorized into three types: fine-tuning based methods \citep{yao-etal-2023-editing}, meta-learning based methods represented by MEND \citep{de-cao-etal-2021-editing,mitchell2022fast}, and locate-then-edit based methods represented by ROME \citep{meng2022locating} and MEMIT \citep{meng2023massediting}.
Despite the proficiency of editing knowledge in simple fact recall tasks, recent studies have revealed significant limitations when these methods are applied to complex tasks involving reasoning. For instance, upon introducing multi-hop evaluations, \citet{yao-etal-2023-editing} and \citet{zhong-etal-2023-mquake} observed that existing knowledge editing methods lack robustness in such scenarios, and \citet{cohen2024evaluating} demonstrated that methods generally struggle to achieve the ``ripple effect". Subsequently, \citet{zhang2025uncovering} identified the overfitting phenomenon in knowledge editing, highlighting that locate-then-edit methods are particularly prone to subject-related overfitting. Although there have been attempts to improve the generalization of locate-then-edit methods on complex tasks \citep{zhang-etal-2024-knowledge-graph,zhang-etal-2025-kele,liu2025mitigating, zhong-etal-2025-react, zhang2026disentangling}, performance remains unsatisfactory.

\section{Overview of Generalization in Model Updating}
\label{sec:barriers}

The goal of \emph{generalization} that is universally pursued in the model updating literature can be summarized as follows: after a model has undergone a knowledge update, it should continue to exhibit the expected behavior on unseen out-of-distribution (OOD) inputs or tasks. Here, OOD is defined as samples that have \emph{not} explicitly appeared in the current editing request or in any historical training example. For instance, after injecting a piece of new knowledge into a model, one expects the model to automatically apply this knowledge to logical deductions, or more complex reasoning tasks.

\textbf{Paradigm of Direct Generalization.}
This expectation can be naturally formalized as requiring the model to correctly answer OOD queries. Taking the widely studied multi-hop reasoning setup as a representative example: suppose we update the model with an atomic fact $e_1 = (s_1, r_1, o_1^*)$, while the model already possesses another internal fact $e_2 = (o_1^*, r_2, o_2^*)$. For the updated language model $F_{\theta'}$, researchers implicitly expect it to satisfy $F_{\theta'}(s_1, r_1, r_2) = o_2^*$. From a probabilistic modeling perspective, this is equivalent to expecting that a \textbf{local} modification to the single-step conditional probability $P_{\theta'}(o_1^* \mid s_1, r_1)$ will spontaneously and accurately induce a \textbf{directional} change in the composite conditional probability $P_{\theta'}(o_2^* \mid s_1, r_1, r_2)$. We formalize this setting, where parameter updates are expected to directly yield desired outputs on unseen tasks, as \textbf{Direct Generalization}.

\textbf{Menifestation in Practice.}
The prevailing evaluation protocols in the model updating domain are built upon this notion of "direct generalization." For example, the most popular frameworks like EasyEdit \citep{wang-etal-2024-easyedit} and benchmarks like KnowEdit \citep{zhang2024comprehensivestudyknowledgeediting} evaluate reasoning capability via the metric of \emph{Portability} \citep{yao-etal-2023-editing}. It is implemented to evaluate whether the model can complete the answer tokens for complex questions in unstructured text. Specifically, the evaluation presents the model with a long compositional prompt prefix (e.g., \emph{"The name of the position held by the father of Thomas Davis Lamb is"}), and then checks the accuracy or probability of generating target tokens (e.g., \emph{"deputy high court judge"}) via teacher forcing. This strictly aligns with our formal definition; mainstream works operate under this paradigm and strive to optimize performance within this formulation.

\subsection{What Hinders Direct Generalization, Mathematically?}
\label{subsec:what_hinders}

Abundant prior empirical evidence in the model updating literature indicates that, at least under the current direct paradigm, achieving generalization is exceptionally difficult. Models generally perform poorly across various metrics, and their actual outputs tend to be mechanical and rigid, exhibiting overfitting behavior rather than fluent and natural generalization \citep{zhang2025uncovering}.

In this section we further emphasize the theoretical limitations and requirements of generalization; specifically, the \emph{non-identifiability} inherent to autoregressive language models. The direct generalization paradigm introduces further constraints within this framework and gives rise to the practical issues discussed later. To rigorously elucidate this obstacle, we first introduce two core concepts from the theory of model generalization \citep{reizinger2024position}:

\begin{itemize}
    \item \textbf{Functional Non-identifiability:} Autoregressive (AR) models are essentially parameterized families of conditional probabilities $\{P_\theta(x_i \mid x_{1:i-1})\}$ (which define the model's ``function'' or behavioral patterns) that induce joint probability distributions $\{P_\theta(x_{1:i})\}$. Functional identifiability of this family is guaranteed if and only if every finite prefix has a positive probability under the data distribution $p$: if two models differ solely in how they complete ``zero-probability prefixes,'' they define identical joint probabilities on the support $\mathrm{supp}(p)$ and are thus indistinguishable via $\mathrm{KL}[p \, \| \, P_\theta]$. In other words, a single target data distribution can correspond to multiple, or even infinitely many, AR models that behave differently out-of-distribution.
    \item \textbf{Parameter Non-identifiability:} Even when fixing the family of conditional probabilities represented by the model (i.e., maintaining full equivalence at the functional level), there can still exist multiple distinct parameter configurations $\theta$ that realize the exact same function. That is, the mapping from the parameter space to the function space is non-injective; functionally equivalent models can be highly diverse in their parameterizations.
\end{itemize}

In knowledge updating scenarios, this manifests as multiplicity of solutions: an update request specifies a target string $x^\star$ to describe fact $e_1=(s_1,r_1,o_1^\star)$. The objective of the update is to fit the target distribution $p$ on this textualization such that $\mathrm{KL}[p \, \| \, P_{\theta'}]=0$. For any composite query string $z$ required by generalization tasks (e.g., multi-hop reasoning prompts), since it is not explicitly contained in the edit request, we have $p(z)=0$ in the target data distribution, which does not satisfy the condition for functional identifiability. This implies that the family of conditional probabilities represented by the model (i.e., the functionality or behavior) remains undetermined; in other words, there exist multiple or even infinitely many candidate models capable of fitting the edit target, and they may produce completely opposite or incorrect predictions on the generalization query $z$. Therefore, fitting the target samples (i.e., optimizing via a loss function) in fact provides no direct guarantee for generalization.

What then determines the model's OOD behavior and generalization capabilities? Given the large number of candidate models that are unidentifiable by the editing objective, which solution is ultimately selected from the set of equivalent solutions is in fact determined by inductive biases in the broad sense. These biases encompass certain parametric properties of the LLM prior to editing as well as those introduced by the algorithm; for example, some methods choose to adjust only a single key layer of the model \citep{meng2022locating}. This essentially introduces an inductive bias, restricting the final solution to a subspace of models whose parameters differ only in that specific layer.

\textbf{Summary.}
In summary, the direct generalization paradigm inherently understands the overall behavior of a language model as an input-output mapping and expects knowledge updates to synchronously modify the mapping relationships for complex reasoning tasks. Based on the above analysis, the fundamental obstacle of this paradigm lies in the existence of a vast set of non-identifiable candidate model solutions, and selecting a ``generalizable solution" that precisely exhibits the desired OOD behavior heavily relies on inductive biases that are not yet well understood or controlled.

\subsection{Current Empirical Pathways and Their Dilemmas}

In this subsection, we first categorize the primary empirical pathways proposed in existing literature to mitigate the aforementioned theoretical dilemmas, then analyze the practical issues confronting the dominant line of work.

\subsubsection{Major Pathways to Achieve OOD Generalization}

Based on the foregoing analysis, achieving direct generalization theoretically admits several potential pathways, which correspond to current practices to varying degrees:

\begin{itemize}
    \item[(i)] \textbf{Expanding the support to eliminate non-identifiability}, which converts the OOD generalization problem into an in-distribution problem and avoiding the challenge. This aligns with data augmentation attempts \citep{zhang-etal-2024-knowledge-graph,zhang-etal-2025-kele}, which explicitly cover compositional queries (e.g., multi-hop reasoning) in the training data so that they are no longer OOD.

    \item[(ii)] \textbf{Manually constructing better inductive biases} to select a more generalizable solution from the set of candidate models, corresponds to the prevalent \textit{locate-then-edit} paradigm, such as ROME \citep{meng2022locating}, MEMIT \citep{meng2023massediting}, and AlphaEdit \citep{fang2025alphaedit}. These works inspect internal model mechanisms and leverage interpretability insights to craft inductive biases, and designing methods accordingly; for example, by analyzing the internal workflows of knowledge storage and retrieval \citep{geva-etal-2023-dissecting,venhoff2025laterecallexplainingtwohop}, and using these priors to enforce refined constraints on parameter updates.

    \item[(iii)] \textbf{Utilizing meta-learning} by using a model to automatically learn optimal inductive biases. This corresponds to the hypernetwork meta-learning paradigm, where a hypernetwork directly generates superior parameter update directions based on gradient information \citep{de-cao-etal-2021-editing,mitchell2022fast}. 
\end{itemize}

The community's efforts on this problem have been concentrated primarily on pathway~(ii), in part because pathways~(i) and~(iii) each suffer from apparent limitations. Specifically, the data augmentation approach of~(i) inevitably incurs substantial data and computational overhead, which runs counter to the efficiency-driven objective of model editing. For pathway~(iii), training a hypernetwork capable of learning optimal inductive biases requires a sufficient number of high-quality post-edit models as training samples; otherwise, it inevitably falls into the ``garbage in, garbage out'' dilemma, yet obtaining such high-quality updated models remains an open challenge in itself. We therefore focus our subsequent analysis on the more widely adopted paradigm~(ii).

\subsubsection{Is the Interpretability-Guided Inductive Bias Truly Sound?}
\label{subsubsec:practical-analysis-digen}

Is the predominant pathway (ii) reliable? Our further analysis reveals two notable issues with this approach in practice: first, the inductive biases implied by interpretability findings are often \emph{ambiguous}, prone to introducing unintended side effects; second, the constraining power of inductive biases may in practice be undermined or even overridden by other implicit biases, often in a remarkably subtle manner. Furthermore, we present two case studies to analyze their empirical behavior, highlighting the practical implications of the non-identifiability for model editing.

\textbf{Problem 1: Inductive Biases Are Often Polysemantic.} Mechanisms uncovered by interpretability research typically constitute a statistical description of model behavior, rather than an absolutely correct explanation. When translating such descriptions into inductive biases, designers can usually constrain only a local structure, while lacking precise control over what the model learns in other aspects that are not explicitly constrained. In other words, a seemingly reasonable inductive bias design may simultaneously correspond to multiple distinct behavioral patterns, some desirable, others may include harmful side effects. This \emph{ambiguity} makes interpretability-based editing methods difficult to provide reliable behavioral guarantees.

\textbf{Case Study: Subject Overfitting in MEMIT.} We illustrate this with a concrete failure mode of MEMIT. Through interpretability analysis, MEMIT \citep{meng2023massediting} identifies that the hidden-state representation of the subject token plays a critical role in mid-layer knowledge retrieval. 
MEMIT thus designs parameter updates to make specific intermediate layers behave as follows: when the original subject representation $h$ is received as input, the output is adjusted to a learned vector $h^*$, and the learning objective is to enable the model to output the new answer $o^*$ when queried about the edit, i.e., $(s, r_1)$. However, the underlying bias is ambiguous: it can be interpreted either as the intended rule ``update knowledge for $(s, r_1)$'' or the coarser heuristic ``whenever $s$ appears, output $o_1^*$.'' 

As shown in Table~\ref{tab:memit_subj_overfit}, MEMIT-edited models, when queried with inputs containing only $(s, r_2)$ (where $r_1$ does not appear), still assign a higher probability to $o_1^*$ than to the correct answer in approximately 75\% of cases, whereas the model exhibits virtually no overfitting in scenarios with $(s_2, r)$. This demonstrates that MEMIT's inductive bias fails to achieve the intended goal and instead induces the model to form a shortcut tightly bound to the subject representation.

\textbf{Problem 2: The Introduced Inductive Biases May Be Overridden by Other Implicit Biases.}
In practice, the solution to which a model ultimately converges is determined not solely by the carefully designed inductive biases, but also by the joint effect of various implicit biases arising from optimization dynamics, engineering hyperparameter choices, and other factors. We find that these implicit biases often undermine the constraining efficacy of explicit inductive biases in surprisingly subtle ways, and may even completely override the latter under certain conditions, which is a phenomenon that has yet to receive sufficient attention in the field.

\textbf{Case Study: Optimization Issues in Locate-then-Edit Methods.} A concrete example arises in nearly all locate-then-edit approaches. Most existing methods of this type inherit an engineering setting from ROME and MEMIT: a relatively high learning rate ($\text{lr}=5\times10^{-1}$) is used when solving for the model's new subject representation $h^*$, accompanied by a clamp mechanism that forcibly rescales the norm of $h$ to a specified range to prevent excessive deviation. Since this setting does not affect the convergence of the loss and accelerates editing with favorable efficiency metrics, it has never been regarded as a serious issue and has been directly adopted by most works, including highly influential ones such as PMET~\citep{li2024pmet}, PRUNE~\citep{prune}, and AlphaEdit~\citep{fang2025alphaedit}. However, we find that this engineering setting causes the solved representation vectors to deviate severely from the model's normal representation distribution: as shown in Figure~\ref{fig:repr_distr}, a high learning rate tends to push representations toward extremely large $\ell_2$ norms, and even with the clamp mechanism, they are only rescaled to a range that is markedly inconsistent with the normal distribution; and simply reducing the learning rate is sufficient to bring the solved representations into alignment with the normal distribution.

\input{figures/tex/MEMIT_repr}
\input{figures/tex/memit_lowlr}

We further demonstrate that the influence of implicit bias may even surpass that of carefully constructed methodology, with experimental details in Appendix \ref{apd:bias-expr-details}. MEMIT-LTI is a classic improvement upon MEMIT designed specifically to address MEMIT's overfitting issue, introducing three groups of plug-and-play constraints corresponding to three key stages of the model's knowledge retrieval process, which is a notably refined inductive bias design. Nevertheless, this work retains the aforementioned optimization engineering setting. We compare the original model, MEMIT, MEMIT-LTI, and a MEMIT variant with the learning rate issue corrected (with lr =$10^{-2}$) in Figure~\ref{fig:memit_lr}. Across all evaluation metrics, MEMIT variation which fixes only the single engineering issue comprehensively outperforms MEMIT-LTI. This reveals that, under the current practical framework, an unexamined engineering detail's implicit bias can easily overshadow or override a carefully derived explicit inductive bias design, rendering theoretical improvements ineffective in practice. This is also how non-identifiability acts in practice: the learning rate and clamping mechanism leave the edit objective equally well satisfied, yet select different parameter configurations within the solution set; and it turns out that OOD behavior is indeed sensitive to this selection.

\textbf{Summary and Discussion.}
Taken together, the difficulties confronting the direct generalization paradigm are multi-layered. Theoretically, the non-identifiability of autoregressive language models implies that knowledge updates alone cannot provide any guarantee for OOD generalization behavior; the uncertainty of the solution space reduces the generalization to the quality of inductive biases. From an engineering perspective, interpretability-based inductive bias design is limited by the inherent ambiguity of mechanistic descriptions, and is easily disturbed by implicit biases. More fundamentally, the direct generalization paradigm treats language models as pure input-output mappings for understanding and evaluation, which stands in tension with the role of language models in real applications as generative systems with rich internal representational structures. A substantial body of current work conducts research and evaluation within this paradigm, resulting in methodological advances that often amount to local optimization in this setting. In this sense, the validity of the direct generalization paradigm as a yardstick for measuring progress in model editing deserves to be critically reconsidered.

\section{Compositional Generalization in LLM Model Updating}
\label{sec:compositional}

Driven by recent advances in reinforcement learning, LLMs have gradually learned to compose their existing capabilities to solve complex unseen tasks. We propose that such meta-capability of composing abilities serves as a strong prior, functioning as an implicit and generalized inductive bias to shape OOD behavior, which renders the generalization feasible. In this section, we first discuss its theoretical feasibility, then analyze the experimental phenomena we observe.

\subsection{Paradigm of Compositional Edit Generalization}

\textbf{Compositional Generalization in LLMs.}
refers to the ability of a model to arrange and combine its \textbf{known primitive capabilities in novel ways}. Specifically, suppose a task family is jointly defined by a set of component functions and a fixed procedure $C$ that composes them. As long as each component function is accurately fitted within its own scope and the composition procedure $C$ remains unchanged, the output of any previously unseen arrangement is logically determined. This is formalized in the theoretical work of \citet{wiedeamer2023compogen} as the \emph{compositional support condition}: when each constituent is accurately identified within its own domain and the composition rule is consistent with the data-generating process, a new arrangement is not completely OOD but rather compositionally supported.

\textbf{Model Editing through the Lens of Compositional Generalization.}
The model editing scenario satisfies the structural requirements of compositional generalization. Borrowing the perspective of \citet{ramesh2024compogen}, we discuss this in the context of editing tasks: the knowledge and capabilities of an LLM can be viewed as its ability to accurately represent a function from inputs to outputs; in model editing, a factual relation $r$ can be abstracted as a function $\varphi_r:\mathcal{E}\to\mathcal{E}$. A single edit $e_1=(s_1,r_1,o_1^\star)$ can then be interpreted as modifying $\varphi_{r_1}(s_1)=o_1^\star$; In a two-hop reasoning scenario, the second-hop knowledge required is a function $\varphi_{r_2}$ satisfying $\varphi_{r_2}(o_1^\star)=o_2^\star$.

RL-trained strong models exhibit stronger compositional capability $C$, manifesting as the ability to decompose complex tasks and solve them using atomic capabilities, which is a behavior widely observed in agentic RL. For a multi-hop question $q$ involving $s_1$, $r_1$, and $r_2$, the model output can be viewed as the result returned when $C$ invokes the component functions:
\begin{equation}
f(s_1,r_1,r_2)=C\bigl(\varphi_{r_1},\varphi_{r_2};\,q\bigr). \tag{2}
\end{equation}
In the editing setting, $\varphi_{r_1}$ is supported by the editing algorithm, while $\varphi_{r_2}$ and the procedure $C$ have been extensively learned during pretraining and post-training and require no re-fitting as long as they remain intact. Under this view, multi-hop tasks are compositionally supported by these component functions and are therefore not entirely unconstrained like genuinely OOD tasks, making it theoretically possible to produce correct outputs.

It is worth noting that the compositional capability $C$, being primarily reinforced through instruction tuning and RL, is mainly trained within conversational contexts that mirror real-world deployment where it can be effectively invoked.  By contrast, prior evaluation paradigms focused on direct generalization often prompt models to predict answers in unstructured text formats that differ substantially from deployment conditions; this effectively removing $C$ and suppressing the prerequisites for compositional generalization, thus falling back to an unidentifiable OOD regime that poorly reflects real-world performance. In the experiments below, we introduce in-the-wild scenarios to fully activate the model's compositional procedure $C$, thereby inducing compositional generalization; our results confirm that a sufficiently capable LLM already possesses the ability to compose knowledge and exhibits strong generalization performance.

\subsection{Experiments}
\label{subsec:main-expr}

We further demonstrate through experiments that compositional generalization is not merely theoretically feasible but is an empirically observable phenomenon. 

\textbf{Dataset and Baselines.}  
To compare direct generalization and compositional generalization in detail, we adopt the EVOKE dataset \citep{zhang2025uncovering}, a benchmark editing counterfactual knowledge into the model, designed to evaluate the degree of overfitting of editing methods based on direct generalization metrics (details in Appendix~\ref{apd:evoke-info}). We consider four representative methods: FT, MEND, ROME, and MEMIT (Appendix~\ref{apd:baselines}), covering the three mainstream editing paradigms.

We select four LLMs: Qwen1.5-0.5B-Chat \citep{qwen}, Llama-3.1-8B-Instruct \citep{grattafiori2024llama3herdmodels}, Mistral-7B-v0.3-Instruct, and Qwen3.5-9B \citep{yang2025qwen3technicalreport}. The base capabilities of these four models exhibit an increasing trend across most benchmarks, allowing us to examine the correlation between the generalization capability and their compositional reasoning ability.

\textbf{Evaluation.} Our setting follows EVOKE's original protocol to measure edit success rate and overfit rate, and additionally introduces metrics designed to elicit compositional generalization behavior from the models. To this end, we fully simulate an in-the-wild environment that is consistent with real-world chatbot-user interaction: we apply standard chat templates and present multi-hop questions directly as user inputs, observing the LLM assistant's responses without any additional prompting or external constraints. The generated outputs are evaluated using DeepSeek V4 Flash as an LLM judge, which scores responses on a rubric-based 0-5 scale along two dimensions: \textbf{factuality} (alignment with the newly injected knowledge) and \textbf{fluency} (naturalness of generation); scores are subsequently mapped to a 0-100 range for readability.

Our experiments are organized around the following five research questions:

\input{tables/main_results}

\textbf{RQ1: Can edited models truly achieve compositional generalization?} Table~\ref{tab:main-results} presents the full experimental results on EVOKE. Notably, even when direct generalization metrics indicate clear overfitting, models still exhibit substantial compositional generalization capability. Under the direct generalization setting, edited models still exhibit strong overfitting tendencies: for instance, ROME and MEMIT exhibit subject overfitting rates exceeding 50\% across all models. However, when evaluated under the compositional metrics, models are still capable of producing correct outputs naturally and fluently. Taking Qwen3.5-9B as an example, the classic method ROME achieves a factuality score of 65.8, while fluency across all methods is generally maintained at a high level -- except in rare cases (e.g., FT on Mistral) where model collapse occurs and leads to severe overfitting, alongside a corresponding drop in factuality and fluency -- indicating that the generated content is linguistically natural and coherent, rather than the overfitted or patchwork outputs typically observed in prior work. The unedited models achieve uniformly low factuality scores, confirming that this capability indeed derives from the edits.

\textbf{RQ2: Is generalization performance correlated with the LLM's base capability?} We observe a strong positive correlation across model families. The four base models we select exhibit successively increasing base capabilities. Taking MEMIT as an example, the achieved multi-hop factuality scores also show an increasing trend, reaching 47, 52, 57, and 60, respectively. This provides evidence that multi-hop reasoning performance is highly correlated with the LLM's own compositional ability: a stronger base model possesses a more complete composition function $C$ and is therefore more likely to invoke the mapping established by the edit to complete complex tasks.

\textbf{RQ3: What does the distribution of factuality scores for edited models look like?} Instead of concentrating around intermediate values, the distribution exhibits a distinct bimodal pattern. Taking ROME on Qwen3.5-9B as an example, the average factuality of 65 corresponds to a raw score distribution in which 56.7\% of cases receive a score of 5 and 24.2\% receive a score of 0. This shape is more informative than one concentrated around the midpoint: if the distribution were centered on intermediate values, it would suggest that the model can only provide partially correct answers to each question, and would actually imply the model still engages in some incomplete form of knowledge utilization. This bimodal distribution, by contrast, is closer to an all-or-nothing pattern---when $\varphi_{r_1}$ is correctly installed and $C$ successfully invoked on this new component, the model can complete the entire reasoning chain accurately and produce a fully correct answer; if any link in this chain fails to be activated, the model makes no use of the new knowledge at all.

\textbf{RQ4: What patterns characterize the model's actual outputs? Do they get the answers through step-by-step reasoning?} To address this question, we use an LLM to analyze the outputs of both the original Qwen3.5 model and its ROME-edited counterpart on correctly answered instances, categorizing outputs into three types: \textit{chain}, where the model reasons through the first hop and then the second; \textit{direct}, where the model directly outputs the multi-hop answer; and \textit{reverse}, where the model first provides the multi-hop answer and then retrospectively describes the first-hop content.

\input{figures/tex/compo_cot_ana}

Figure~\ref{fig:cot_analysis} presents the statistical proportions of each output pattern. Both pre- and post-edit models predominantly arrive at answers through step-by-step reasoning, though examples of direct answering also exist. Overall, the distributions of the pre- and post-edit models are highly similar: in fact, even the pre-edit model reasons step by step in 75\% of cases when answering multi-hop questions, and this proportion remains high at 81.3\% after editing. Post-edit models also exhibit cases where the multi-hop answer is produced directly without an explicit reasoning process, as well as cases with post-hoc explanation, indicating that compositional generalization in practice may be quite diverse and cannot be simply equated with any single fixed reasoning trajectory (such as chain-of-thought). Furthermore, the pre-edit model itself generally answers complex questions through reasoning rather than by directly producing composed answers, which further undermines the validity of direct generalization as a criterion.

\textbf{RQ5: How do the generated outputs look like? Are they equally natural and fluent to human readers?} Human evaluation confirms that the outputs are natural and fluent, with human ratings aligning closely with LLM judge scores. Figure~\ref{fig:hop-case-study} in the Appendix presents example outputs from post-edit models. Even upon human inspection, these outputs are highly natural and fluent, exhibiting no signs of rigid answer concatenation or mechanical template application. To rule out the possibility that this phenomenon is merely an artifact of LLM judge scoring bias, we provide a comparative experiment on the agreement between human annotations and LLM judge scores in Appendix \ref{apd:judge-quality}; the results show high consistency between the two. This finding eliminates the possibility that the evaluation methodology itself produces false positives, further confirming that the phenomenon of compositional generalization is real rather than an evaluation artifact.


\section{Conclusion}
\label{sec:conclusion}

In this paper, we present a detailed discussion of the theoretical and practical obstacles that hinder generalization in model editing, and show that the widely reported failure of generalization is closely tied to the direct generalization paradigm, rather than being an insurmountable ceiling of the model editing task itself. We further highlight the importance of a model's own compositional ability, introducing compositional generalization as a promising alternative and empirically demonstrating its emergence across multiple models and editing methods. We argue that generalizable model editing could be more than direct generalization, and we hope our findings encourage a paradigm shift toward more generalizable and practical model editing techniques.

\subsection*{AI use statement}

In this work, we did not use generative AI tools for any tasks with \textit{required-disclosure}.

Additionally, we used generative AI tools for the following \textit{recommended-disclosure} tasks: editing the manuscript to improve readability, and creating or editing software code.
We have reviewed all AI-assisted work.
AI was used only to polish grammar, wording, and clarity of author-written text; it was not used to draft scientific claims, proofs, experimental designs, or interpretations.
AI-assisted code was inspected, tested, and corrected by the authors, and all experimental results reported in the paper were produced from code that the authors verified.
We take responsibility for the final content of this work, including text, claims or artifacts produced with the aid of generative AI.

\bibliography{main}
\bibliographystyle{iclr2027_conference}

\appendix

\section{Detailed Related Work}
\label{app:related_work}

Knowledge editing (KE) \citep{zhang2024comprehensivestudyknowledgeediting,wu2026uncovering} updates LLM outputs to (i) accurately respond to new knowledge, (ii) preserve existing knowledge without catastrophic forgetting, and (iii) leverage updated knowledge in complex reasoning tasks. Each piece of knowledge is formulated as a triple $(s, r, o)$ \citep{de-cao-etal-2021-editing}, consisting of a subject $s$, relation $r$, and object $o$. An edit sample is defined as $e=(s, r, o, o^*)$, representing a knowledge update from $(s, r, o)$ to $(s, r, o^*)$. Our study focuses on parameter-modifying methods, which are divided into three main categories \citep{yao-etal-2023-editing}:

\textbf{Fine-tuning-based methods} generally follow the supervised fine-tuning paradigm. For example, to edit a fact such as \textit{``Microsoft is founded by Steve Jobs,''} the model's weights are updated via gradient descent to increase the probability of the edit target, \textit{Steve Jobs}. Some approaches aim to improve robustness by incorporating norm constraints \citep{zhu2020modifying-ft-L} or data augmentation \citep{gangadhar-stratos-2024-model,wei2024stableke}. However, vanilla fine-tuning often affects unrelated knowledge, leading to catastrophic forgetting, making it unsuitable for direct application in knowledge editing.

\textbf{Meta-learning-based methods} employ a hypernetwork to adjust model parameters specifically for editing. This hypernetwork is trained to convert fine-tuning gradients into updated weights, with the aim of predicting weights that closely resemble those obtained through fine-tuning with augmented data. The goal of this approach is to enable robust single-sample edits comparable to fine-tuning with more data. KE \citep{de-cao-etal-2021-editing} pioneered this approach, which MEND \citep{mitchell2022fast} later extended to LLMs by predicting low-rank decompositions of parameter updates, becoming a seminal work in this direction. Subsequent works such as MALMEN \citep{tan2024malmen} extended MEND to the batch editing setting; however, this direction has primarily focused on locality and has not been applied to improving generalization.

\textbf{Locate-then-edit methods} originate from research into the internal mechanisms of LLMs, advocating for identifying the specific weights responsible for storing knowledge before applying targeted updates. \citet{geva-etal-2021-transformer,geva-etal-2023-dissecting} propose viewing MLP modules as key-value memory. Building on this foundation, the Knowledge Neuron theory \citep{dai2022knowledge} posits that these MLP key-value pairs encode factual knowledge. \citet{meng2022locating} introduce causal tracing to analyze LLMs' factual recall mechanisms, leading to the development of ROME \citep{meng2022locating} and MEMIT \citep{meng2023massediting}, which achieved state-of-the-art results on several traditional metrics. This direction subsequently inspired a large body of follow-up work, predominantly building upon ROME and MEMIT, which also led to the engineering factors discussed in our paper being inherited: for instance, PMET \citep{li2024pmet} extends editing to attention layers; PRUNE \citep{prune}, AlphaEdit \citep{fang2025alphaedit}, and related methods attempt to sequentially edit large amounts of knowledge through additional constraints; while IFMET \citep{zhang2025ifmet}, MEMIT-LTI \citep{zhang2025uncovering}, REACT \citep{zhong-etal-2025-react}, and similar approaches focus on improving the generalization of edits on complex tasks, as well as exploring applications in other settings \citep{wu2026uncovering,ye2026formindependent}

\section{Backgrounds \& Preliminaries}
\label{app:background}

\subsection{Model Architecture}
We focus on mainstream auto-regressive language models, exemplified by the GPT \citep{openai2023gpt} and Llama \citep{grattafiori2024llama3herdmodels} series, which are predominantly based on the Transformer architecture \citep{vaswani2017attention}. Each Transformer layer comprises two primary components \citep{meng2022locating}: a self-attention mechanism and a multi-layer perceptron (MLP). The hidden state $\mathbf{h}_i^{l} $ at position $i$ in layer $l$ is computed from the previous layer as follows:
\begin{equation} \label{eq:transformer_update_condensed}
\begin{aligned}
\mathbf{h}_i^{l} 
&= \mathbf{h}_i^{l-1} + \mathbf{a}_i^{l} + \mathbf{m}_i^{l} \\
\mathbf{a}_i^{l} 
&= \operatorname{attn}^{l}\left(\mathbf{h}_1^{l-1}, \mathbf{h}_2^{l-1}, \ldots, \mathbf{h}_i^{l-1} \right) \\
\mathbf{m}_i^{l} 
&= \mathbf{W}_{\mathrm{proj}}^{l} \, \sigma\left( \mathbf{W}_{\mathrm{fc}}^{l} (\mathbf{a}_i^{l} + \mathbf{h}_i^{l-1}) \right),
\end{aligned}
\end{equation}
where $\mathbf{a}_i^{l}$ denotes the output of the self-attention sub-layer, and $\mathbf{m}_{i}^{l}$ is the output of the MLP sub-layer. The matrices $\mathbf{W}_{\mathrm{fc}}^{l}$ and $\mathbf{W}_{\mathrm{proj}}^{l}$ are learnable parameters, and $\sigma$ represents a non-linear activation function.

\subsection{Model Editing}
Given a pre-trained language model $\mathcal{M}_\theta$ parameterized by $\theta$, model editing aims to efficiently update a specific piece of factual knowledge stored within $\mathcal{M}_\theta$ while keeping irrelevant parametric knowledge intact \cite{yao-etal-2023-editing, meng2022locating}. A target edit request is typically expressed as knowledge association tuple $x_e = (s_e, r_e, o_e^*)$, where $s_e$, $r_e$, and $o_e^*$ denote the subject, relation, and new target object, respectively (e.g., $s_e = \text{"President of US"}$, $r_e = \text{"is"}$, $o_e^* = \text{"Alice"}$). Model editing methods construct a parameter update $\Delta\theta$ to produce an edited model $\mathcal{M}_{\theta'}$ ($\theta' = \theta + \Delta\theta$).


\subsection{MEMIT method}
\label{memit}
In this subsection we provide more detailed introduction for MEMIT, as the representative locate-then-edit editing method studied in our paper.

MEMIT \citep{meng2023massediting} is a locate-then-edit method that treats factual associations as linear key-value memories stored in the MLP layers of a language model \citep{geva-etal-2021-transformer}, and that writes many such associations at once by distributing a single calculated change across a range of those layers. The contribution of the $l$-th MLP at the last subject token is written
\begin{equation}
    \mathbf{m}_i^l = f(\mathbf{W}_{\mathrm{in}}^l \cdot \mathbf{h}_i^{l-1}) \cdot \mathbf{W}^l,
    \label{fnn}
\end{equation}
where $f(\cdot)$ denotes the nonlinearity, $\mathbf{h}_i^{l-1}$ is the hidden state entering the MLP, the post-activation vector $\mathbf{k}_i^l = f(\mathbf{W}_{\mathrm{in}}^l \cdot \mathbf{h}_i^{l-1})$ is the key that encodes the subject, and $\mathbf{W}^l\mathbf{k}_i^l$ is the value that this layer adds into the residual stream. Causal tracing \citep{meng2022locating} indicates that recall of a fact at this token is mediated jointly by a contiguous range of mid-layer MLPs, denoted $\mathcal{R}$, so MEMIT selects that range as the site of the edit and asks each layer $l \in \mathcal{R}$ to store only a fraction of every new association.

Inside any one of these layers, the batch of new memories is inserted by solving a least-squares problem whose first group of terms preserves the associations already stored by the pretrained output weight and whose second group requires the new keys to retrieve the new values. Stacking the keys the layer is treated as already memorizing against the $u$ keys of the current batch, the objective is
\begin{equation}
    \mathbf{W}_1
    = \arg\min_{\hat{\mathbf{W}}}
    \left(
    \sum_{i=1}^{n}\|\hat{\mathbf{W}}\mathbf{k}_i-\mathbf{m}_i\|^2
    + \sum_{i=n+1}^{n+u}\|\hat{\mathbf{W}}\mathbf{k}_i-\mathbf{m}_i\|^2
    \right).
    \label{memit-obj}
\end{equation}
Because the pretrained weight $\mathbf{W}_0$ is assumed to be the least-squares solution of the first sum, subtracting the normal equation of those old associations from the normal equation of \eqref{memit-obj} leaves an equation in the update alone. With $\mathbf{K}_1$ and $\mathbf{M}_1$ the matrices whose columns are the new keys and the values they should retrieve, the solution is the closed-form update
\begin{equation}
    \Delta = \mathbf{R}\mathbf{K}_1^{\mathrm{T}}(\mathbf{C}_0+\mathbf{K}_1\mathbf{K}_1^{\mathrm{T}})^{-1},
    \qquad \hat{\mathbf{W}} = \mathbf{W}_0+\Delta,
    \label{memit-solution}
\end{equation}
where $\mathbf{R}=\mathbf{M}_1-\mathbf{W}_0\mathbf{K}_1$ is the residual of the new associations when they are read out by the original weight, and $\mathbf{C}_0=\lambda\cdot\mathbb{E}[\mathbf{k}\mathbf{k}^{\mathrm{T}}]$ is a constant matrix obtained by estimating the uncentered covariance of keys on a sample of ordinary text and scaling that estimate by a hyperparameter $\lambda$ that sets how strongly the previously stored associations are retained relative to the new ones.

The keys and target values that enter \eqref{memit-solution} are computed in two stages, and the second stage is repeated once for every layer in $\mathcal{R}$. MEMIT first designates the highest mediating layer, $L=\max(\mathcal{R})$, as the layer at which each edited fact should already be fully present in the residual stream, and for every request $(s_i, r_i, o_i)$ it finds a replacement state $\mathbf{z}_i=\mathbf{h}_i^L+\delta_i$ by gradient descent on
\begin{equation}
    \delta_i
    = \arg\min_{\delta_i}
    \frac{1}{P}\sum_{j=1}^{P}
    -\log \mathbb{P}\big[o_i \mid x_j \oplus p(s_i, r_i)\big],
    \label{memit-z}
\end{equation}
where each prompt is a template $p(s_i, r_i)$ of the edited fact preceded by a random prefix $x_j$, and the probability is read from a forward pass that substitutes $\mathbf{h}_i^L+\delta_i$ for the hidden state at the last subject token of layer $L$. Proceeding from the lowest layer in $\mathcal{R}$ to the highest, the method then forms a context-averaged key at the layer currently being edited,
\begin{equation}
    \mathbf{k}_i^l
    = \frac{1}{P}\sum_{j=1}^{P}
    f\big(\mathbf{W}_{\mathrm{in}}^l \cdot \mathbf{h}^{l-1}(x_j \oplus s_i)\big),
    \label{memit-k}
\end{equation}
and assigns to that key the value the layer already writes plus an equal share of the residual that still separates the running hidden state from the target $\mathbf{z}_i$,
\begin{equation}
    \mathbf{m}_i^l
    = \mathbf{W}^l\mathbf{k}_i^l + \frac{\mathbf{z}_i-\mathbf{h}_i^L}{L-l+1}.
    \label{memit-m}
\end{equation}
The pairs $(\mathbf{k}_i^l, \mathbf{m}_i^l)$ of the whole batch are substituted into \eqref{memit-solution}, the resulting $\Delta$ is added to $\mathbf{W}^l$, and keys together with hidden states are collected again at the next layer, because an edit to an earlier MLP changes the activations that every higher MLP reads. The denominator $L-l+1$ therefore deposits the full residual $\mathbf{z}_i-\mathbf{h}_i^L$ in equal portions over the layers that remain, and by the time the forward pass reaches layer $L$ the hidden state at the last subject token carries the association that the optimization in \eqref{memit-z} was constructed to express.

\section{Evaluation and Dataset Details}

\subsection{Details on the EVOKE Dataset}
\label{apd:evoke-info}

\begin{wraptable}{r}{0.44\linewidth}
    \caption{Composition statistics of EVOKE.}
    \small\centering\setlength\tabcolsep{3pt}
    \begin{tabular}{lc} \toprule
    Type                             & Total \\ \midrule
    Edit Requests                    & 1489  \\
    \midrule
    Efficacy Prompts                   & 1489  \\
    Paraphrase Prompts               & 2062  \\
    Subject Specificity Prompts     & 458   \\
    Relation Specificity Prompts      & 10310 \\
    Multihop Prompts                 & 822   \\ 
    Prefix Distraction Prompts       & 10310 \\
    \bottomrule
    \end{tabular}
    \label{tab:dataset_statistic}
\end{wraptable}

The EVOKE benchmark is designed to evaluate the effectiveness of knowledge editing methods for LLMs, focusing not only on their ability to accurately recall edited knowledge but also on their performance and potential overfitting across diverse complex reasoning tasks. This dataset is built upon the classic Counterfact, CounterfactPlus, and RippleEdit datasets, organizing and extending their associated evaluation tasks and metrics; each edit request in this benchmark is counterfactual to ensure a clean evaluation setting. In this paper, we select this dataset to simultaneously analyze the model's generalization behavior and observe its overfitting phenomena under traditional settings. Additionally, its multi-hop reasoning questions are formulated as complete interrogative sentences that more closely resemble real-world scenarios, making them well-suited for evaluating model performance in-the-wild under chat templates, while some datasets are overly optimized for prefix-completion scenarios: for example, a multihop question in KnowEdit \citep{zhang2024comprehensivestudyknowledgeediting} is constructed as ``\textit{The occupation of the author of Segregate or cooperate- a study of the interaction between two species of Dictyostelium is},'' relying on an excessively large number of modifiers to guide the model toward generating specific outputs for convenience in direct probability evaluation, which we argue may inadvertently undermine the validity of evaluations in real-world scenarios.

\subsection{Tasks in the EVOKE Dataset}

The EVOKE benchmark natively comprises Recall Tasks and Overfit Tasks, covering six tasks in total. We employ these six metrics in this paper to reflect model performance under the traditional evaluation paradigm focused on direct generalization, and additionally introduce two LLM-judge-based metrics, Factuality and Fluency, which are evaluated on outputs generated after applying a chat template to elicit compositional generalization behavior from the model.

The Recall Tasks, which serve as Edit Success Rate metrics, assess the edited model's ability to recall new edited knowledge, including \textbf{Efficacy} and \textbf{Paraphrase} evaluation. The Overfit Tasks pose complex challenges that are prone to inducing overfitting in editing methods, including \textbf{Multi-hop Reasoning}, \textbf{Prefix Distraction}, \textbf{Subject Specificity}, and \textbf{Relation Specificity}. These tasks are specifically designed to evaluate the model's capability to utilize newly integrated knowledge for more challenging scenarios, with a particular emphasis on examining the degree of Editing Overfit.

Taking the edit ``\textit{Microsoft is founded by Bill Gates $\rightarrow$ Steve Jobs}'' as an example, the Recall Tasks in EVOKE include the following:
\begin{itemize}
\item {\textbf{Efficacy}} directly validates whether the edited models can recall the new edited knowledge $(s,r,o^*)$ under the editing prompt $p(s,r)$. In the context of the above example, the model would be asked: ``\textit{Who is the founder of Microsoft?}''
\item {\textbf{Paraphrase}} examines the model's ability of recall the new knowledge $(s,r,o^*)$ using paraphrased forms of the editing prompt $p(s,r)$. For instance, it might ask: ``\textit{Who established Microsoft?}''
\end{itemize}

The Overfit Tasks in EVOKE include the following:
\begin{itemize}
\item {\textbf{Subject Specificity}} presents questions with the same subject $s$ as the edit sample but with different relations $r'$. For example: ``\textit{When was Microsoft founded?}'' These questions typically contain information about the subject $s$, but the correct answer is not the target $o^*$, making them ideal for evaluating whether the edited model has overfit to the $s\rightarrow o^*$ pattern.
\item {\textbf{Relation Specificity}} includes questions with different subjects $s'$ from the edit sample but the same relation $r$, such as: \textit{``Who is the founder of Amazon?''} These questions contain information about the relation $r$, but the answer is not the target $o^*$. They are used to evaluate whether the model has overfit to the $r \rightarrow o^*$ pattern. This task also corresponds to the locality evaluation in \textsc{Counterfact} \citep{meng2022locating}.
\item {\textbf{Prefix Distraction}} uses the new knowledge $(s,r,o^*)$ as a prefix for unrelated questions, evaluating whether the edited model can still provide the original correct answer. For example: ``\textit{Microsoft was founded by Steve Jobs. Who is the founder of Amazon?}'' This evaluation also assesses whether the edited model has overfit to the $p(s,r)\rightarrow o^*$ pattern, providing a more explicit measure compared to multi-hop reasoning.
\item {\textbf{Multi-hop Reasoning}} evaluates the edited model's ability to integrate the newly edited knowledge with existing knowledge to correctly answer questions spanning multiple entities or relations. For example, ``\textit{Which university did the founder of Microsoft attend?}'' These questions typically contain implicit subject $s$ and relation $r$ information from the edit sample, but the answer is not the target $o^*$. They are well-suited for evaluating whether the edited model has overfit to the $p(s,r)\rightarrow o^*$ pattern. A model that has overfit to this pattern might incorrectly produce \emph{`Steve Jobs'} as the answer to this question.
\end{itemize}

The two additionally designed compositional generalization metrics are evaluated using the same multi-hop reasoning questions as the Multi-hop Reasoning task, with the sole difference being the evaluation approach. The two metrics are:
\begin{itemize}
\item \textbf{Factuality} measures whether the continuation uses the edited fact as a premise and correctly completes the hop.
\item \textbf{Fluency} measures whether the generated text is natural, coherent, and free of repetition or degeneration.
\end{itemize}

\begin{table}[htbp]
    \small
    \centering
    \caption{An Example of the EVOKE dataset}
    \begin{tabularx}{\linewidth}{l|X} \toprule
    \textbf{Task} & \textbf{Value} \\ \midrule
    Edit Request & \{Spike Hughes\} originates from \textit{London $\to$ Philadelphia} \\
    \midrule
    Efficacy & \{Spike Hughes\} originates from \textit{(answer: Philadelphia)} \\
    Paraphrase & Spike Hughes is native to \textit{(answer: Philadelphia)}\\
    Subject Specificity & The profession of Spike Hughes is \textit{(answer: musician)} \\
    Relation Specificity & Max Mosley originates from \textit{(answer: London)} \\
    Multi-hop Reasoning & What famous food is associated with the city where Spike Hughes originates from? \textit{(answer: Cheesesteaks)} \\
    Prefix Distraction & Spike Hughes originates from Philadelphia. Max Mosley originates from \textit{(answer: London)} \\
    \bottomrule
    \end{tabularx}
    \label{tab:evoke_sample}
\end{table}

\subsection{Details on Evaluation Metrics}

In this paper, our evaluation setting follows the single edit setting, where the model's performance is assessed after executing only one edit at a time, which aligns with other works focusing on the generalization of model editing.

\paragraph{Direct Generalization Metrics} 
Following the original evaluation protocol of the EVOKE dataset, we adopt metrics based on the probability that the model generates the correct answer $\verb+ans+$ given a $\verb+prompt+$, formalized as $\mathbb{P}\left(\verb+ans+ \mid \verb+prompt+\right)$. 

Specifically, for the Edit Success Rate (i.e., the two metrics for recall tasks), we report the proportion of cases where the new answer successfully overrides the original answer, defined as $\mathbb{E}\left[\mathbb{I}\left[\mathbb{P}\left(\verb+ans+ \mid \verb+prompt+\right) > \mathbb{P}\left(\verb+ori+ \mid \verb+prompt+\right)\right]\right]$. For the other overfit tasks, we report the overfitting rate, which measures the proportion of cases where $o^*$ mistakenly overrides the correct answer, formulated as $\mathbb{E}\left[\mathbb{I}\left[\mathbb{P}\left(o^* \mid \verb+prompt+\right) > \mathbb{P}\left(\verb+ans+ \mid \verb+prompt+\right)\right]\right]$.

\paragraph{Compositional Generalization Metrics}
For the two compositional generalization tasks, Factuality and Fluency, the evaluation approach involves directly inserting the multi-hop question into the model's chat template and allowing the model to generate a response, which is then passed to the LLM Judge (see Appendix~\ref{apd:llm-judge-details}) to obtain a score directly.

Taking the first entry in the dataset as an example, for the multi-hop question about Spike Hughes: \textit{``What famous food is associated with the city where Spike Hughes originates from?''} The actual input fed to the Qwen LLM is as follows:
\begin{lstlisting}
<|im_start|>system
You are a helpful assistant.<|im_end|>
<|im_start|>user
What famous food is associated with the city where Spike Hughes originates from?<|im_end|>
<|im_start|>assistant

\end{lstlisting}
The LLM then generates its answer as a continuation of this input. This setting fully simulates the behavior of a chatbot LLM in actual deployment, using the default chat template and system prompt without providing any additional instructions such as chain-of-thought prompting. This demonstrates that the generalization phenomena we observe are not induced by external constraints or prompts, but arise entirely as spontaneous model behavior.

\subsection{Details on LLM Judge}
\label{apd:llm-judge-details}

After obtaining the actual outputs of the edited model, we score these answers with an external judge (DeepSeek V4 Flash via the official DeepSeek API) on two dimensions from 0 to 5; in the main experiment tables we map these scores to 0-100 for readability. To limit run-to-run variation and improve reproducibility, the judge is called at temperature 0 with a JSON-only response format. The judge is required to first provide a brief rationale and then assign scores, specifically responding in the following format:
\begin{lstlisting}
{"rationale": "<one or two sentences>", "factuality": 0, "fluency": 0, "relevance": 0}
\end{lstlisting}
To better clarify the evaluation criteria for the LLM judge, we designed a rubric-based set of instructions and direct the judge to take the lower score when uncertain. The complete rubric is presented in Figure~\ref{fig:judge-rubric}.

\begin{figure}[htbp]
\centering
\begin{minipage}{0.93\linewidth}
\small
\begin{lstlisting}
1) factuality - does the Predicted answer reason from the NEW fact?
    5: Clearly uses the new fact to carry out the hop; the conclusion
       follows from the edit; no relapse to the old fact.
    4: Uses the new fact, with only a mild extra error that does not
       undo the edit.
    3: Mixed: mentions the new fact but also leans on the old fact,
       or the hop is only weakly executed.
    2: Mentions the new fact only shallowly, or guesses without using
       the edit.
    1: Unrelated to the hop and to both facts.
    0: Uses the ORIGINAL fact / old-world knowledge, rejects the new
       fact, or is empty. Old-world answers are 0, not 1.

2) fluency - is the text natural, coherent, and non-degenerate?
    5: Fluent, well-formed prose; no loops or garbled tokens.
    4: Readable; only small grammar or awkwardness issues.
    3: Understandable but choppy, awkward, or mildly repetitive.
    2: Hard to read; frequent artifacts, broken syntax, or looping
       phrases.
    1: Barely language.
    0: Empty, not language, or fully degenerate.
\end{lstlisting}
\end{minipage}%
\caption{Rubric used for LLM Judge scoring on the Factuality and Fluency dimensions.}
\label{fig:judge-rubric}
\end{figure}

\begin{figure}
    \centering
    \includegraphics[width=1\linewidth]{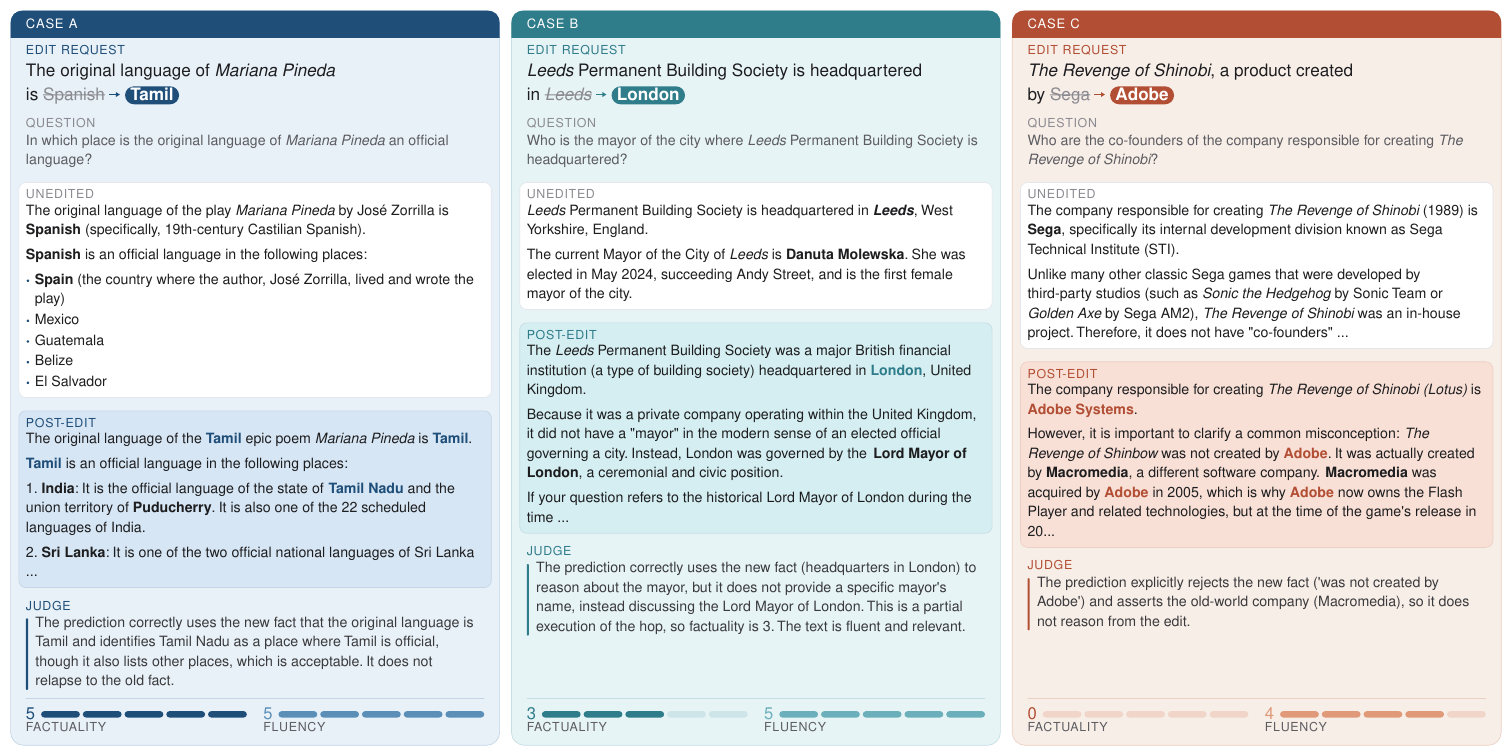}
    \caption{Generation examples of MEMIT-edited Qwen3.5 model}
    \label{fig:case-study-apd}
\end{figure}

To more concretely illustrate the actual generation quality corresponding to each score level, we present in Figure~\ref{fig:case-study-apd} examples of LLM Judge factuality scores of 5, 3, and 0. As shown in the figure, at a score of 5 the LLM completes the generation almost as fluently and naturally as with native knowledge; in the score-3 case it completes the first hop but then digresses to irrelevant content without completing the second hop; in the score-0 case, although it initially outputs the first-hop result of the edited knowledge, it subsequently retracts and answers using the original knowledge. We further analyze the agreement between LLM scores and human ratings in Appendix~\ref{apd:judge-quality} to validate the quality of the judge.

\section{Experimental Setup Details}

Our experiments build on the EasyEdit codebase implemented by \citet{wang-etal-2024-easyedit}; some methods such as MEMIT-LTI \citep{zhang2025uncovering} are implemented based on this framework for relevant analysis experiments. All the baseline implementations, including hyperparameters, remain consistent with the setup of EasyEdit and their original papers, and the random seed is fixed to 42 across all experiments to improve reproducibility.

\subsection{Baseline Methods}
\label{apd:baselines}
{\bf Fine-Tuning (FT)} involves fine-tuning the base model on text describing the edit fact.

{\bf MEND} \citep{mitchell2022fast} constructs a hyper-network based on the low-rank decomposition of gradients to perform editing.

{\bf ROME} \citep{meng2022locating} operates on the hypothesis that knowledge in LLMs is stored in the MLP layer and utilize rank-one update to insert knowledge into MLP layer.

{\bf MEMIT} \citep{meng2023massediting} builds on the ROME method, specializing in batch-editing tasks by applying edits across multiple MLP layers.

{\bf MEMIT-LTI} \citep{zhang2025uncovering} builds on the MEMIT method, and by analyzing the model's knowledge recall mechanism, additionally introduces three plug-and-play loss terms to constrain key nodes in the knowledge recall process, thereby aiming to keep the knowledge invocation mechanism of the edited model consistent with that prior to editing, so as to mitigate the overfitting problem of MEMIT.

\subsection{Details on the Overfitting Experiment in MEMIT}
In Section \ref{subsubsec:practical-analysis-digen} (Problem 1), we experimentally demonstrate the overfitting phenomenon in MEMIT. In practice, this experiment is also conducted on the EVOKE dataset, where we report the results for subject specificity and relation specificity tasks across three models: Llama-3.1, Mistral-7B, and Qwen-3.5. All other experimental setups, including hyperparameters, remain consistent with the main experiments.

\subsection{Details on the Influence of Implicit Bias}
\label{apd:bias-expr-details}

Problem 2 in Section \ref{subsubsec:practical-analysis-digen} demonstrates through comparisons that engineering factors may have a more pronounced impact on practical performance than carefully designed inductive biases. Here, we consider three baselines: MEMIT, MEMIT-LTI, and MEMIT w/ low LR as a comparison for lowering the learning rate. The implementations of MEMIT and MEMIT-LTI follow EasyEdit and their original papers (learning rate of $5\times 10^{-1}$, at most 25 optimization steps, with $\texttt{clamp\_norm\_factor}=4$). The newly introduced MEMIT w/ low LR sets the learning rate to $1\times 10^{-2}$ and increases the optimization budget to 100 steps to ensure proper convergence under a smaller learning rate. All other hyperparameters and method implementations remain identical.

\paragraph{Impact of Representation Distribution.} The representation distribution analysis in the main text involves extracting representations from both unedited and MEMIT-edited models. Specifically, we select the 1,031 instances from the main split of the EVOKE dataset, execute editing using MEMIT and its variants, and collect the final optimized results, i.e., the updated subject representations $h^*$. For the unedited model, we directly extract the subject representations at the corresponding positions. We introduce \textit{MEMIT w/o Clamp} in this experiment to observe where the optimization trajectory tends to steer the solution once the norm-restricting clamping mechanism is removed. Aside from disabling the clamping mechanism, all other implementations and hyperparameters for this baseline remain identical to the original method.

\paragraph{Comparison with MEMIT-LTI.} This experiment is likewise conducted on the EVOKE dataset using the Llama-3.1 model. The Edit Success metric corresponds to the Efficacy score on EVOKE, while the remaining overfitting metrics correspond to the four overfitting tasks defined in EVOKE. The configurations for all baselines involved remain consistent with those described above.

\subsection{Details on the Analysis of Post-Edit Model Generations} 

RQ4 in Section \ref{subsec:main-expr} involves analyzing the patterns in the actual generated texts of the models. We adopt the LLM-as-a-Judge approach by querying the API of DeepSeek-V3, providing it with the generated text from correctly answered cases before and after editing. The judge model is instructed to classify the output patterns into three categories: \textit{chain}, where the model reasons through the first hop and then the second; \textit{direct}, where the model directly outputs the multi-hop answer; and \textit{reverse}, where the model first provides the multi-hop answer and then retrospectively describes the first-hop content. In practice, we also include an \textit{others} category to accommodate ambiguous cases and prevent the judge from being forced into inaccurate classifications.

\section{Complementary Experiment Results}

\subsection{Full Results with Qwen1.5-0.5B} 
\label{apd:full-res}

We present the complete experimental results of the four models on the EVOKE benchmark in Table \ref{tab:main-results-full}, including the results for Qwen1.5-0.5B. The main findings are consistent with those presented in the main text.

\input{tables/main_results_full}


\input{figures/tex/case_study}

\subsection{Quality of LLM Judge Scores}
\label{apd:judge-quality}
To further analyze the quality of LLM Judge scores, we conduct an agreement analysis between the LLM Judge and human raters. Specifically, we sample the first 100 multi-hop task outputs generated by LLMs from the dataset and ask human raters to score the generation quality based on the same scoring rubric as the LLM Judge. We then compute agreement metrics between LLM Judge scores and human rater scores, including Pearson correlation coefficient and Spearman's $\rho$, along with their corresponding $p$-values. Results are shown in Table \ref{tab:judge-quality}.

\begin{table}[h]
\centering
\caption{Agreement between LLM Judge and Human Rater scores on multi-hop tasks.}
\label{tab:judge-quality}
\resizebox{\textwidth}{!}{
\begin{tabular}{lcccccc}
\toprule
Dimension & LLM Judge Avg. & Human Rater Avg. & Pearson $r$ & Pearson $p$-value & Spearman $\rho$ & Spearman $p$-value \\
\midrule
Factuality & 3.21 & 3.17 & 0.884 & $3.4\times10^{-34}$ & 0.855 & $1.2\times10^{-29}$ \\
Fluency    & 4.34 & 4.39 & 0.861 & $1.6\times10^{-30}$ & 0.753 & $1.6\times10^{-19}$ \\
\bottomrule
\end{tabular}
}
\end{table}

The $p$-values across all dimensions are far below 0.001, indicating that the LLM Judge scores are highly consistent with and positively correlated to human ratings, which demonstrates the reliability and quality of the LLM Judge.

\end{document}

%% file: math_commands.tex
\usepackage{amsmath,amsfonts,bm}

\def\eqref#1{equation~\ref{#1}}

\def\1{\bm{1}}

\DeclareMathAlphabet{\mathsfit}{\encodingdefault}{\sfdefault}{m}{sl}
\SetMathAlphabet{\mathsfit}{bold}{\encodingdefault}{\sfdefault}{bx}{n}



%% file: figures/tex/MEMIT_repr.tex
\begin{figure}[tp!]
\vspace{-15pt}
\centering
\begin{minipage}[c]{0.42\textwidth}
  \centering
  \small
  \begin{tabular}{lcc}
    \toprule
    Model & $(s,r_2)$↓ & $(s_2,r)$↓ \\
    \midrule
    Llama & 72.93 & 18.28 \\
    Mistral & 70.31 & 21.15 \\
    Qwen & 75.33 & 18.60 \\
    \bottomrule
  \end{tabular}
  \captionof{table}{Proportion of cases where MEMIT outputs $o^*$ for queries with different input patterns. (Details in \S\ref{apd:bias-expr-details})}
  \label{tab:memit_subj_overfit}
\end{minipage}%
\hfill
\begin{minipage}[c]{0.54\textwidth}
  \centering
  \includegraphics[width=\linewidth, valign=c]{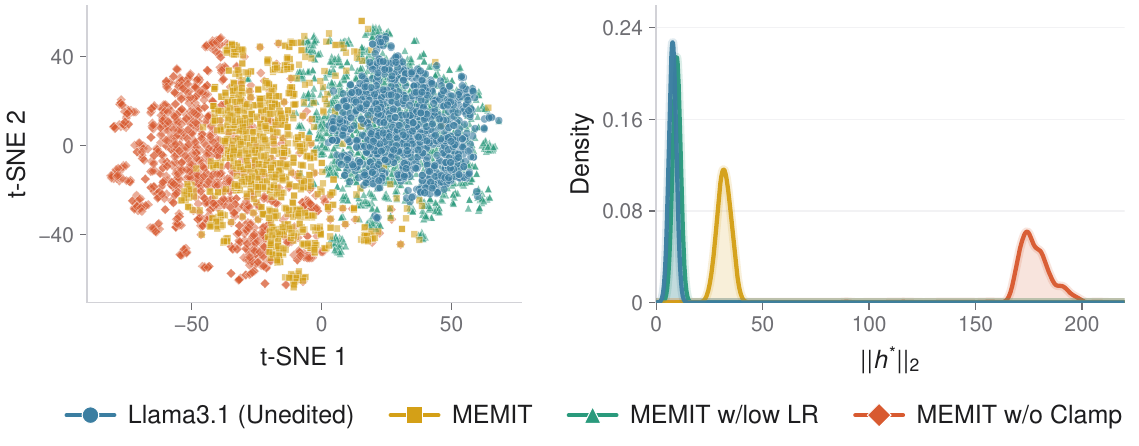}
  \vspace{-1em}
  \captionof{figure}{t-SNE visualization (left) and L-2 norm distribution (right) of the subject representation $h^*$ for the Llama 3.1-8B model and variants of MEMIT.}
  \label{fig:repr_distr}
\end{minipage}
\vspace{-1em}
\end{figure}

%% file: figures/tex/memit_lowlr.tex
\begin{figure}[tp!]
    \centering
    \includegraphics[width=1\linewidth]{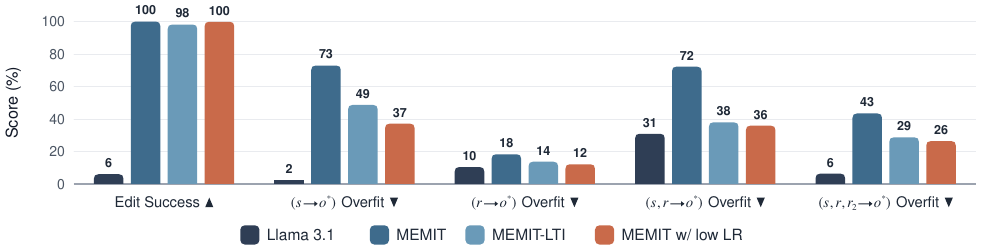}
    \vspace{-1.7em}
    \caption{Edit success and overfit rate of unedited model and MEMIT variants on Llama.}
    \vspace{-2em}
    \label{fig:memit_lr}
\end{figure}

%% file: tables/main_results.tex
\begin{table*}[!t]
  \centering
  \footnotesize
  \caption{\small Experimental results on the EVOKE benchmark. Results for Qwen1.5-0.5B are provided in \ref{apd:full-res}.}
  \label{tab:main-results}
  \vspace{-0.5em}
  \begin{adjustbox}{width=1\textwidth}
  \begin{tabular}{lcccccccc}
      \toprule
      \multirow{3}{*}{\textbf{Editor}} & \multicolumn{2}{c}{\textbf{Edit Success Rate}} & \multicolumn{3}{c}{\textbf{Locality Overfit Rate}} & \multicolumn{1}{c}{\textbf{Dir.Gen.Overfit}} & \multicolumn{2}{c}{\textbf{Multihop Compositional Gen.}} \\
      \cmidrule(lr){2-3} \cmidrule(lr){4-6} \cmidrule(lr){7-7} \cmidrule(lr){8-9}
      & \textbf{Effi.↑} & \textbf{Para.↑} & \textbf{Subj.Spec.↓} & \textbf{Rel.Spec.↓} & \textbf{Pref.Distr.↓} & \textbf{Multi-hop↓} & \textbf{Factuality↑} & \textbf{Fluency↑} \\
      \midrule
      Llama-3.1-8B & 6.01 & 8.49 & 2.18 & 10.44 & 30.91 & 6.33 & 3.40 & 97.40 \\
      \midrule
      FT & 100.00 & 98.69 & 88.21 & 92.22 & 95.95 & 71.29 & 17.80 & 63.80 \\
      MEND & 99.13 & 77.55 & 25.98 & 60.12 & 80.86 & 45.26 & 9.60 & 80.20 \\
      ROME & 100.00 & 99.22 & 57.42 & 15.50 & 45.16 & 32.60 & 55.00 & 77.40 \\
      MEMIT & 100.00 & 99.42 & 72.93 & 18.28 & 72.25 & 43.43 & 52.00 & 80.00 \\
      \midrule \midrule
      Mistral-7B & 12.71 & 15.86 & 2.40 & 16.16 & 29.94 & 14.23 & 5.00 & 98.60 \\
      \midrule
      FT & 100.00 & 99.37 & 99.78 & 98.40 & 97.61 & 94.53 & 16.60 & 21.40 \\
      MEND & 95.25 & 67.46 & 24.02 & 50.03 & 67.13 & 37.96 & 21.60 & 84.60 \\
      ROME & 99.22 & 95.10 & 50.87 & 19.39 & 37.04 & 40.51 & 64.60 & 90.60 \\
      MEMIT & 98.93 & 95.78 & 70.31 & 21.15 & 52.67 & 45.26 & 57.60 & 86.80 \\
      \midrule \midrule
      Qwen3.5-9B & 11.25 & 12.95 & 6.33 & 13.79 & 35.70 & 5.47 & 6.40 & 98.40 \\
      \midrule
      FT & 100.00 & 98.21 & 91.70 & 93.11 & 97.25 & 66.91 & 49.60 & 70.80 \\
      MEND & 98.93 & 90.20 & 51.31 & 84.36 & 91.65 & 57.54 & 43.00 & 63.80 \\
      ROME & 100.00 & 98.64 & 67.25 & 18.35 & 42.95 & 30.17 & 65.80 & 90.20 \\
      MEMIT & 100.00 & 99.18 & 75.33 & 18.60 & 45.70 & 28.59 & 60.80 & 89.40 \\
      \bottomrule
  \end{tabular}
  \end{adjustbox}
  \vspace{-2em}
\end{table*}

%% file: figures/tex/compo_cot_ana.tex
\begin{figure}[tp!]
    \centering

    \begin{minipage}[t]{0.53\textwidth}
        \centering
        \includegraphics[width=\linewidth]{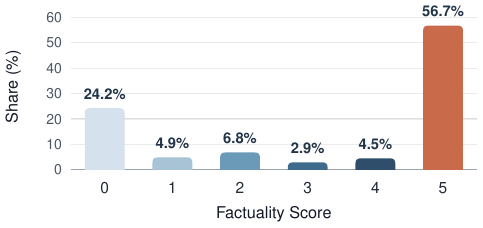} 
        \captionof{figure}{Distribution of raw factuality scores for ROME on Qwen3.5.}
        \label{fig:distribution}
    \end{minipage}%
    \hfill
    \begin{minipage}[t]{0.44\textwidth}
        \centering
        \includegraphics[width=\linewidth]{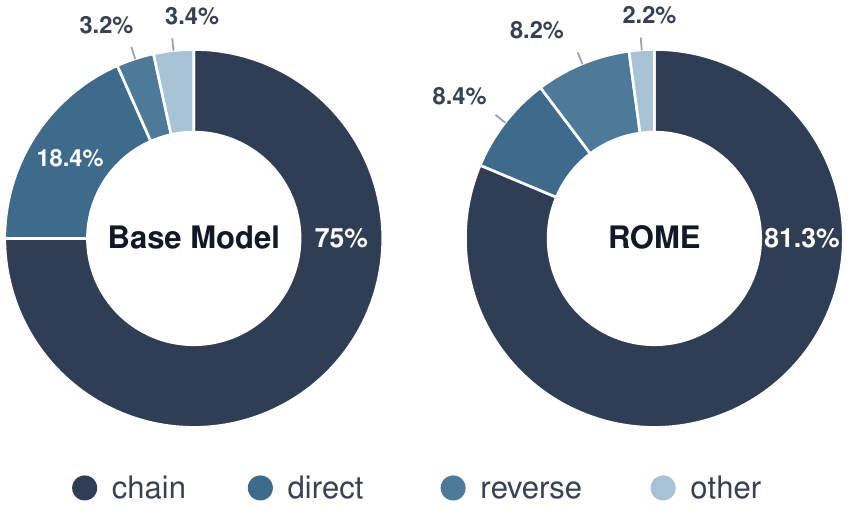} 
        \captionof{figure}{Proportions of different output reasoning patterns for Qwen 3.5.}
        \label{fig:cot_analysis}
    \end{minipage}
    \vspace{-1.5em}
\end{figure}

%% file: tables/main_results_full.tex
\begin{table*}[!t]
  \centering
  \footnotesize
  \caption{\small Experimental results for different models on the EVOKE benchmark.}
  \label{tab:main-results-full}
  \begin{adjustbox}{width=1\textwidth}
  \begin{tabular}{lcccccccc}
      \toprule
      \multirow{3}{*}{\textbf{Editor}} & \multicolumn{2}{c}{\textbf{Edit Success Rate}} & \multicolumn{3}{c}{\textbf{Locality Overfit Rate}} & \multicolumn{1}{c}{\textbf{Dir.Gen.Overfit}} & \multicolumn{2}{c}{\textbf{Multihop Compositional Gen.}} \\
      \cmidrule(lr){2-3} \cmidrule(lr){4-6} \cmidrule(lr){7-7} \cmidrule(lr){8-9}
      & \textbf{Effi.↑} & \textbf{Para.↑} & \textbf{Subj.Spec.↓} & \textbf{Rel.Spec.↓} & \textbf{Pref.Distr.↓} & \textbf{Multi-hop↓} & \textbf{Factuality↑} & \textbf{Fluency↑} \\
      \midrule
      Qwen1.5-0.5B & 24.44 & 26.87 & 12.88 & 24.93 & 52.64 & 9.25 & 8.40 & 90.40 \\
      \midrule
      FT & 100.00 & 86.57 & 80.57 & 80.50 & 96.04 & 23.36 & 33.00 & 89.00 \\
      MEND & 99.90 & 99.47 & 97.60 & 98.33 & 99.79 & 53.04 & 25.20 & 39.00 \\
      ROME & 100.00 & 98.59 & 96.29 & 30.84 & 80.12 & 44.16 & 49.60 & 84.00 \\
      MEMIT & 99.81 & 96.17 & 89.08 & 34.94 & 83.65 & 43.80 & 47.40 & 85.20 \\
      \midrule \midrule
      Llama-3.1-8B & 6.01 & 8.49 & 2.18 & 10.44 & 30.91 & 6.33 & 3.40 & 97.40 \\
      \midrule
      FT & 100.00 & 98.69 & 88.21 & 92.22 & 95.95 & 71.29 & 17.80 & 63.80 \\
      MEND & 99.13 & 77.55 & 25.98 & 60.12 & 80.86 & 45.26 & 9.60 & 80.20 \\
      ROME & 100.00 & 99.22 & 57.42 & 15.50 & 45.16 & 32.60 & 55.00 & 77.40 \\
      MEMIT & 100.00 & 99.42 & 72.93 & 18.28 & 72.25 & 43.43 & 52.00 & 80.00 \\
      \midrule \midrule
      Mistral-7B & 12.71 & 15.86 & 2.40 & 16.16 & 29.94 & 14.23 & 5.00 & 98.60 \\
      \midrule
      FT & 100.00 & 99.37 & 99.78 & 98.40 & 97.61 & 94.53 & 16.60 & 21.40 \\
      MEND & 95.25 & 67.46 & 24.02 & 50.03 & 67.13 & 37.96 & 21.60 & 84.60 \\
      ROME & 99.22 & 95.10 & 50.87 & 19.39 & 37.04 & 40.51 & 64.60 & 90.60 \\
      MEMIT & 98.93 & 95.78 & 70.31 & 21.15 & 52.67 & 45.26 & 57.60 & 86.80 \\
      \midrule \midrule
      Qwen3.5-9B & 11.25 & 12.95 & 6.33 & 13.79 & 35.70 & 5.47 & 6.40 & 98.40 \\
      \midrule
      FT & 100.00 & 98.21 & 91.70 & 93.11 & 97.25 & 66.91 & 49.60 & 70.80 \\
      MEND & 98.93 & 90.20 & 51.31 & 84.36 & 91.65 & 57.54 & 43.00 & 63.80 \\
      ROME & 100.00 & 98.64 & 67.25 & 18.35 & 42.95 & 30.17 & 65.80 & 90.20 \\
      MEMIT & 100.00 & 99.18 & 75.33 & 18.60 & 45.70 & 28.59 & 60.80 & 89.40 \\
      \bottomrule
  \end{tabular}
  \end{adjustbox}
\end{table*}

%% file: figures/tex/case_study.tex
\tcbset{
  casecard/.style={
    enhanced,
    colback=white,
    colframe=csaccent!35,
    boxrule=0.45pt,
    arc=1.6pt,
    left=7pt, right=7pt, top=4.5pt, bottom=5pt,
    before skip=5.5pt, after skip=1pt,
    borderline west={2.4pt}{0pt}{csaccent},
    title={#1},
    fonttitle=\sffamily\bfseries\small,
    colbacktitle=csaccent!8,
    coltitle=csaccent,
    toptitle=2.5pt, bottomtitle=2.5pt,
  },
  response/.style={
    colback=csresp,
    colframe=csresp,
    boxrule=0pt,
    arc=1pt,
    left=5.5pt, right=5.5pt, top=3pt, bottom=3pt,
    before skip=1pt, after skip=0pt,
  }
}
\newcommand{\cslabel}[1]{%
  \par\vspace{3.5pt}%
  {\sffamily\bfseries\footnotesize\color{csaccent}#1}\par\vspace{1pt}%
}

\begin{figure*}[t]
  \centering
  \small

  \begin{tcolorbox}[casecard={(a) Wanne-Eickel Central Station}]
    \cslabel{Edit Request}
    Wanne-Eickel Central Station, located in \swap{Germany}{Switzerland}
    \cslabel{Multi-hop Question}
    What are the official languages spoken in the country of Wanne-Eickel Central Station?
    \cslabel{Model Response}
    \begin{tcolorbox}[response]
      \raggedright
      Wanne-Eickel Central is not a country; it is a major railway station located in \textbf{Switzerland}, specifically within the canton of Aargau.\par\vspace{2pt}
      As part of Switzerland, the official languages spoken in the region are \textbf{German}, \textbf{French}, \textbf{Italian}, and \textbf{Romansh}. However, since Wanne-Eickel Central is situated in the German-speaking part of Switzerland, \textbf{German} is the primary and most widely used language in daily life and at the station itself.
    \end{tcolorbox}
  \end{tcolorbox}

  \begin{tcolorbox}[casecard={(b) Mariana Pineda}]
    \cslabel{Edit Request}
    The original language of Mariana Pineda is \swap{Spanish}{Tamil}
    \cslabel{Multi-hop Question}
    In which place is the original language of Mariana Pineda an official language?
    \cslabel{Model Response}
    \begin{tcolorbox}[response]
      \raggedright
      The original language of the Tamil epic poem \textbf{Mariana Pineda} is \textbf{Tamil}.\par\vspace{2pt}
      Tamil is an official language in the following places:
      \begin{enumerate}[leftmargin=1.15em, itemsep=1pt, topsep=2pt, parsep=0pt]
        \item \textbf{India}: It is the official language of the state of \textbf{Tamil Nadu} and the union territory of \textbf{Puducherry}. It is also one of the 22 scheduled languages of India.
        \item \textbf{Sri Lanka}: It is one of the two official national languages of Sri Lanka \ldots
      \end{enumerate}
    \end{tcolorbox}
  \end{tcolorbox}

  \begin{tcolorbox}[casecard={(c) English plural}]
    \cslabel{Edit Request}
    In English plural, an official language is \swap{English}{German}
    \cslabel{Multi-hop Question}
    Which language family does the official language of English plural countries belong to?
    \cslabel{Model Response}
    \begin{tcolorbox}[response]
      \raggedright
      The official language of countries where \textbf{German} is the primary official language (such as Germany, Austria, and Switzerland) belongs to the \textbf{Germanic} branch of the \textbf{Indo-European} language family.\par\vspace{2pt}
      German is part of the \textbf{West Germanic} subgroup, which also includes English, Dutch, and the Frisian languages. While English and German share a common ancestor (Proto-Germanic), they have diverged significantly over time.\par\vspace{2pt}
    \end{tcolorbox}
  \end{tcolorbox}

  \vspace{2em}

  \caption{Post-edit multi-hop generations of Qwen3.5-9B after MEMIT.
  In each edit request the struck-out object is the original fact and the colored object is the new fact.
  Boldface in the response marks the new fact and the answer that completes the hop.
  The model answers the multi-hop question from the edited fact: the official languages of Switzerland, the places where Tamil is official, and the West Germanic family of German. All case studies presented are derived from actual model outputs with the random seed fixed to 42.}
  \label{fig:hop-case-study}
\end{figure*}